\documentclass[runningheads]{llncs}
\usepackage[T1]{fontenc}
\usepackage{graphicx,verbatim}
\usepackage{booktabs}
\usepackage{xcolor}
\makeatletter
\newcommand{\printfnsymbol}[1]{\textsuperscript{\@fnsymbol{#1}}}
\makeatother
\begin{document}
\title{Patient-Specific Articulated Digital Twins from a Single Full-Body CT Scan}

\author{
Han Zhang\textsuperscript{\(\dagger\)}\thanks{Corresponding author}\inst{1}
\and
Boyang Zhao\textsuperscript{\(\dagger\)}\inst{1}
\and
Mathias Unberath\inst{1}
}

\authorrunning{H. Zhang et al.}

\institute{
\textsuperscript{\(\dagger\)}Equal contribution.\\
Johns Hopkins University, Baltimore MD 21218, USA \\
\email{\{hzhan206, bzhao27, unberath\}@jhu.edu} \\
}

\maketitle

\begin{abstract}
Patient-specific anatomical models provide individualized context for surgical planning, image-guided intervention, and algorithm development. However, most CT-derived models are static: they preserve the body configuration captured at scan time, but cannot represent how the same anatomy would appear after patient repositioning. This limitation is especially important for radiographic imaging, where appearance depends jointly on imaging geometry and patient pose. We present a proof-of-concept for constructing a patient-specific \emph{articulated} digital twin from a single full-body CT scan. The method fits a parametric human body model (SMPL) to obtain a patient-aligned kinematic scaffold, binds segmented bones and organs to an anatomy-aware rig, and retargets body-pose changes while preserving skeletal geometry. Using full-body CT scans from three subjects, the fitted scaffold achieved 15.8 $\pm$ 4.0 mm chamfer distance and 95.9 $\pm$ 1.8\% skeletal enclosure. Recomposition at the acquisition pose preserved major radiographic structure, with overall SSIM of 0.872 $\pm$ 0.016 and PSNR of 18.5 $\pm$ 1.4 dB across paired DRRs. Across unseen target poses, the resulting twins enabled articulation while maintaining high skeletal enclosure (94.4 $\pm$ 0.4\%). As a feasibility demonstration, we render the articulated twin as pose-dependent DRRs. These results suggest the feasibility of extending static, view-controllable CT simulation toward pose-controllable anatomical twins for future synthetic imaging and positioning studies.

\end{abstract}
\section{Introduction}

Patient-specific anatomical modeling from medical imaging is a foundation for surgical planning, image-guided intervention, and medical AI development~\cite{Gao2023SyntheticDA,jcm14113989}. By providing digital representations of individual anatomy, such models enable controlled virtual testing of procedures and algorithms before deployment in real patients. Most existing patient-specific models, however, are static and capture the patient in only the anatomical configuration present during image acquisition~\cite{mekki_digital_2025,shetty_boss_2023}. This is adequate for analyzing anatomy in the acquired pose, but becomes limiting whenever patient configuration changes across imaging protocols, positioning procedures, or interventional workflows~\cite{midtgaard_patient_2023}. A static reconstruction preserves fine anatomical detail, yet says nothing about how that anatomy would appear once the body moves. Bridging this gap requires a patient digital twin that is not only individualized but also articulable into new body configurations.

This limitation is most apparent in radiographic imaging, where the projected appearance is determined jointly by the imaging geometry and the spatial configuration of the patient anatomy. Two families of models offer partial remedies, but neither provides anatomy that is simultaneously patient-specific and posable. Population-level anatomical phantoms such as XCAT~\cite{dahal_xcat-30_2024} support pose and motion variation, but represent anatomy through a fitted template phantom rather than the patient's own geometry. Conversely, CT-derived digitally reconstructed radiographs (DRRs) are widely used for synthetic X-ray generation, 2D/3D registration, and image-guided intervention research~\cite{gopalakrishnan2023intraoperative,killeen_stan_2024,unberath_deepdrr_2018}. These simulations preserve patient-specific CT geometry and attenuation, and can leverage dense CT-derived anatomical labels~\cite{Wasserthal_2023}, but keep the anatomy locked to the pose captured at CT acquisition. CT-based simulation is therefore view-controllable but not pose-controllable: it can move the projection geometry, but cannot place the patient into a new pose.

\begin{figure}[t]
    \includegraphics[width=\textwidth]{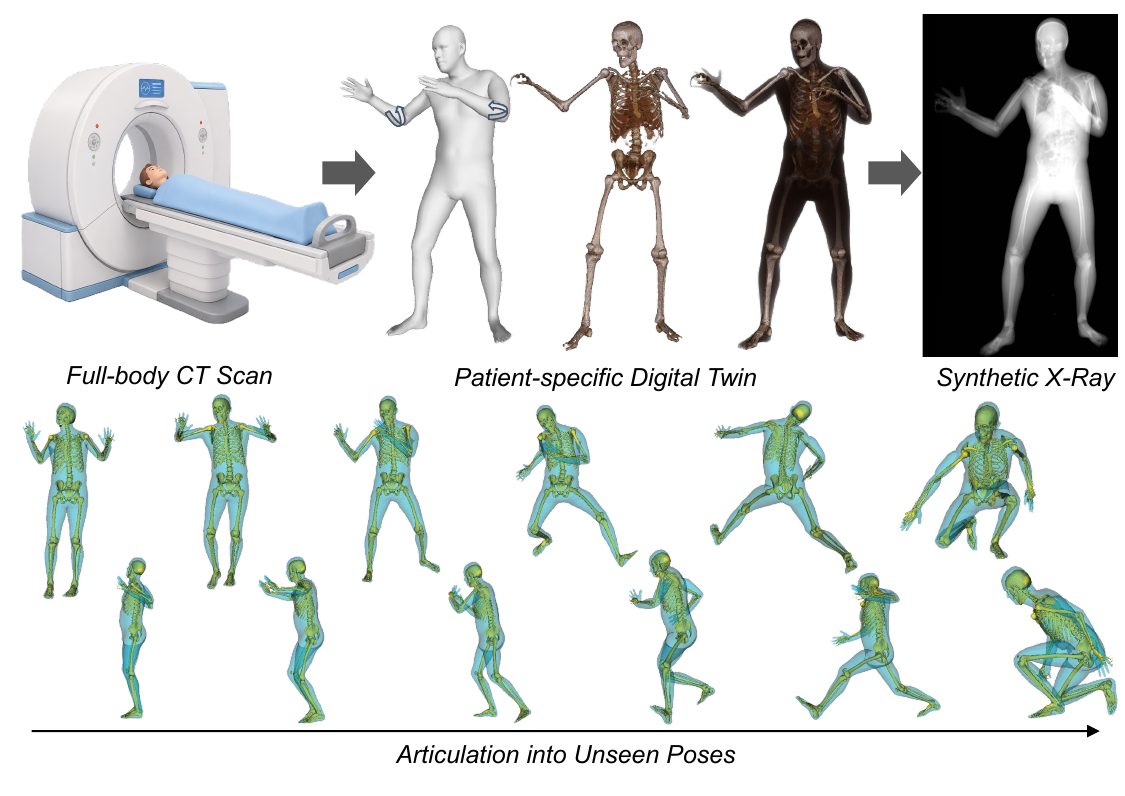}
    \caption{Overview. A single full-body CT scan (top left) is converted into a patient-specific articulated digital twin that fuses CT-derived skeletal anatomy with a fitted body envelope. The twin can be articulated into diverse unseen poses (bottom) and rendered as pose-dependent synthetic radiographs (top right).} \label{teaser}
\end{figure}

In this work, we present a proof-of-concept for constructing a patient-specific articulated digital twin from a single full-body CT scan, as illustrated in Fig.~\ref{teaser}. Here, \emph{patient-specific} refers to the anatomical content of the twin: the skeletal and organ geometry, Hounsfield units, and joint anchors are derived from the individual's CT, while the body envelope and kinematic prior are provided by the population-level SMPL model~\cite{loper_smpl_2023}. Starting from CT-derived skeletal anatomy, we fit a parametric human body model to obtain a patient-aligned kinematic scaffold, and bind the segmented anatomy to a reduced kinematic rig. Then, we retarget body-pose changes onto the patient's anatomy, enabling articulation into different configurations while preserving the reconstructed skeletal geometry as rigid structures. We characterize the twin through its body-fitting accuracy, its recomposition fidelity at the acquisition pose, and its structural self-consistency under articulation. As a downstream feasibility demonstration, we render the articulated anatomy as DRRs, illustrating that a single static CT can be turned into a source of pose-diverse synthetic radiographs. 

\section{Methods}
\begin{figure}[t]
    \includegraphics[width=\textwidth]{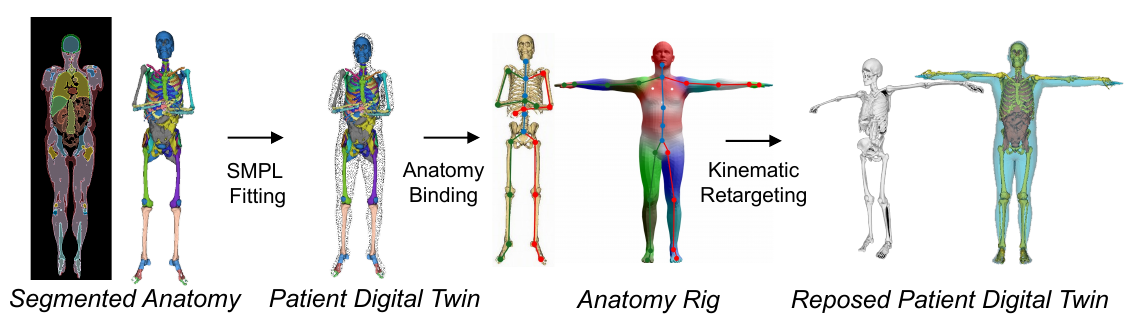}
    \caption{Overview of the pipeline. } \label{pipeline}
\end{figure}

Given a full-body CT scan, our goal is to construct a patient-specific articulated digital twin that can be reposed into new body configurations. As shown in Fig.~\ref{pipeline}, the pipeline consists of three stages: patient anatomical modeling, anatomy-aware kinematic binding, and pose retargeting.

\subsection{Patient anatomical modeling}
The goal of this stage is to construct a static patient digital twin from the CT scan. The digital twin consists of CT-derived anatomical geometry with a fitted SMPL body model~\cite{loper_smpl_2023} that provides an articulated kinematic scaffold. Given a full-body CT volume, we first reconstruct anatomical geometry. Specifically, we use TotalSegmentator~\cite{Wasserthal_2023} to extract semantic segmentations of the body, bones, and organs. We then apply Marching Cubes to each segmentation mask, followed by Taubin smoothing, to obtain surface meshes of the skeleton and corresponding organ structures.

We next fit an SMPL body model to the CT-derived anatomy. The goal of this fitting step is to estimate the global orientation $R$, translation $T$, shape parameters $\beta$, and body pose parameters $\theta$ that best describe the patient's body in the CT acquisition pose. Directly fitting SMPL to the CT-derived body surface can be unreliable because the external surface in a supine CT scan may exhibit substantial soft-tissue deformation due to gravity and contact with the scanner table. We therefore initialize the SMPL model using skeletal keypoint alignment. Specifically, we initialize SMPL using 25 OpenPose-style anatomical keypoints~\cite{8765346} manually annotated on the segmented patient skeleton. These keypoints are used only for initialization, and this step could be replaced by automatic skeletal landmark detection in future work.

Starting from this initialization, we perform multi-phase optimization to refine the SMPL shape and pose parameters. The articulated body model is optimized by minimizing
\[
(R^*,T^*,\beta^*,\theta^*)
=
\arg\min_{R,T,\beta,\theta}
\left(
\lambda_{\mathrm{chamfer}} L_{\mathrm{chamfer}}
+
\lambda_{\mathrm{inclusion}} L_{\mathrm{inclusion}}
+
\lambda_{\mathrm{prior}} L_{\mathrm{prior}}
\right),
\]
where $L_{\mathrm{chamfer}}$ enforces geometric consistency between the SMPL body surface and the CT-derived body surface, $L_{\mathrm{inclusion}}$ encourages the CT-derived skeletal anatomy to remain inside the fitted body envelope, and $L_{\mathrm{prior}}$ regularizes the SMPL shape and pose parameters. The fitting prioritizes enclosing the skeletal anatomy within the body envelope ($L_{\mathrm{inclusion}}$) over tightly matching the body surface ($L_{\mathrm{chamfer}}$), as the SMPL prior is biased toward standing poses whereas full-body CT is acquired supine.

The output of this stage is a static anatomical digital twin in the CT acquisition pose,
\[
\mathcal{D}_A =
\left(
\mathcal{M}_{\mathrm{bone}}^A,
\mathcal{M}_{\mathrm{organ}}^A,
M_{\mathrm{SMPL}}(\beta,\theta,R,T)
\right),
\]
where $\mathcal{M}_{\mathrm{bone}}^A$ and $\mathcal{M}_{\mathrm{organ}}^A$ preserve patient-specific anatomy, and the fitted SMPL model provides the patient-aligned kinematic scaffold used in subsequent articulation.

\subsection{Anatomy-aware kinematic binding}

The goal of this stage is to convert the static anatomical digital twin into a kinematic rig anchored to the patient's own skeletal geometry. We define a reduced kinematic tree rooted at the pelvis (shown in Fig.~\ref{pipeline}) and assign each CT-derived bone to a semantically corresponding joint or joint group. Standard SMPL animation applies linear blend skinning to a surface mesh, blending multiple joint transforms per vertex to produce smooth skin deformation around joints. This is inappropriate for skeletal geometry, where each bone should move as a rigid body rather than deform. Instead, we assign each CT-derived bone to joint groups and apply the corresponding rigid transform, transferring the SMPL kinematic prior to the anatomy while preserving per-bone rigidity. 

Instead of directly using SMPL joint locations, we define CT-specific joint anchors from the segmented patient skeleton, providing patient-specific rotation centers for the reconstructed bones. Specifically, each joint anchor is estimated from the closest surface region between a child bone and its parent bone. Organs and soft-tissue structures are assigned to coarser anatomical segments, including the pelvis, lower lumbar spine, mid lumbar spine, thorax, and skull.

The output of this stage is an anatomy rig,
\[
\mathcal{R}_{\mathrm{CT}}
=
\left(
\mathcal{B}^A,
\mathcal{T}_{\mathrm{CT}},
\mathcal{A}_{\mathrm{bone}},
\mathcal{A}_{\mathrm{organ}},
\mathcal{J}_{\mathrm{CT}}^A
\right),
\]
where $\mathcal{B}^A$ denotes the segmented anatomy from the CT scan, $\mathcal{T}_{\mathrm{CT}}$ is the reduced kinematic tree, $\mathcal{A}_{\mathrm{bone}}$ and $\mathcal{A}_{\mathrm{organ}}$ define bone and organ bindings, and $\mathcal{J}_{\mathrm{CT}}^A$ denotes the CT-specific joint anchors in the acquisition pose.

\subsection{Pose retargeting}
The goal of this stage is to articulate the patient digital twin into a target body configuration. Given the anatomy rig $\mathcal{R}_{\mathrm{CT}}$, we transfer the pose change defined by SMPL to the CT-derived anatomy. Target body poses can be obtained from motion sequences or specified manually as SMPL pose parameters $\theta$.

Let $\theta_A$ denote the fitted SMPL pose corresponding to the CT acquisition pose, and let $\theta_B$ denote a target body pose. The relative joint motion from $\theta_A$ to $\theta_B$ is then extracted from the SMPL kinematic chain and propagated through the reduced CT kinematic tree using forward kinematics. For each CT-derived anatomical structure, we apply the rigid transformation of its assigned joint or anatomical segment. Bones are transformed according to their associated joint anchors. Organs and soft-tissue structures are transformed at the segment level, allowing internal anatomy to follow the motion of the surrounding skeletal structures.  For example, the heart and lungs move with the thorax.

The output of this stage is a reposed patient digital twin,
\[
\mathcal{D}_B =
\left(
\mathcal{M}_{\mathrm{bone}}^B,
\mathcal{M}_{\mathrm{organ}}^B,
\mathcal{M}_{\mathrm{SMPL}}^B
\right),
\]
where $\mathcal{M}_{\mathrm{bone}}^B$, $\mathcal{M}_{\mathrm{organ}}^B$, and $\mathcal{M}_{\mathrm{SMPL}}^B$ denote the reposed skeletal, organ, and body envelope geometries under the target pose $B$. This produces patient anatomy in previously unseen body configurations, carrying over the skeletal structure reconstructed from the original CT scan as rigid bodies, while organ and soft-tissue geometry follows the motion of its assigned anatomical segment.

\section{Experiments and Results}
As a proof-of-concept, we evaluate the proposed approach using full-body CT scans from three subjects in the NMDID dataset~\cite{edgar2020nmdid}. The original CT scans were segmented using TotalSegmentator~\cite{Wasserthal_2023}, and converted into surface meshes for fitting, kinematic binding, and pose retargeting. To illustrate pose-dependent radiographic simulation, we render the articulated twin as DRRs using DeepDRR~\cite{unberath_deepdrr_2018}, a CT-based X-ray simulation framework widely used for synthetic radiograph generation and image-guided intervention research~\cite{killeen_fluorosam_2026,killeen_pelphix_2023}. Since DeepDRR operates on volumes, the twin is voxelized into a labeled CT volume before rendering. Each anatomical component retains its source-CT Hounsfield units (HU): bones, organs, and soft-tissue segments are repositioned via their assigned rigid transforms, while the residual volume inside the reposed SMPL body envelope is filled with a constant mean HU value derived from muscle tissue in the original CT. We quantitatively assess the twin from three perspectives: geometric fitting accuracy, DRR recomposition fidelity at the acquisition pose, and structural self-consistency under articulation. All reported results are presented as descriptive proof-of-concept measurements rather than as estimates of population-level performance across variations in body habitus, pathology, or scanning protocol.

\begin{figure}[t]
    \includegraphics[width=\textwidth]{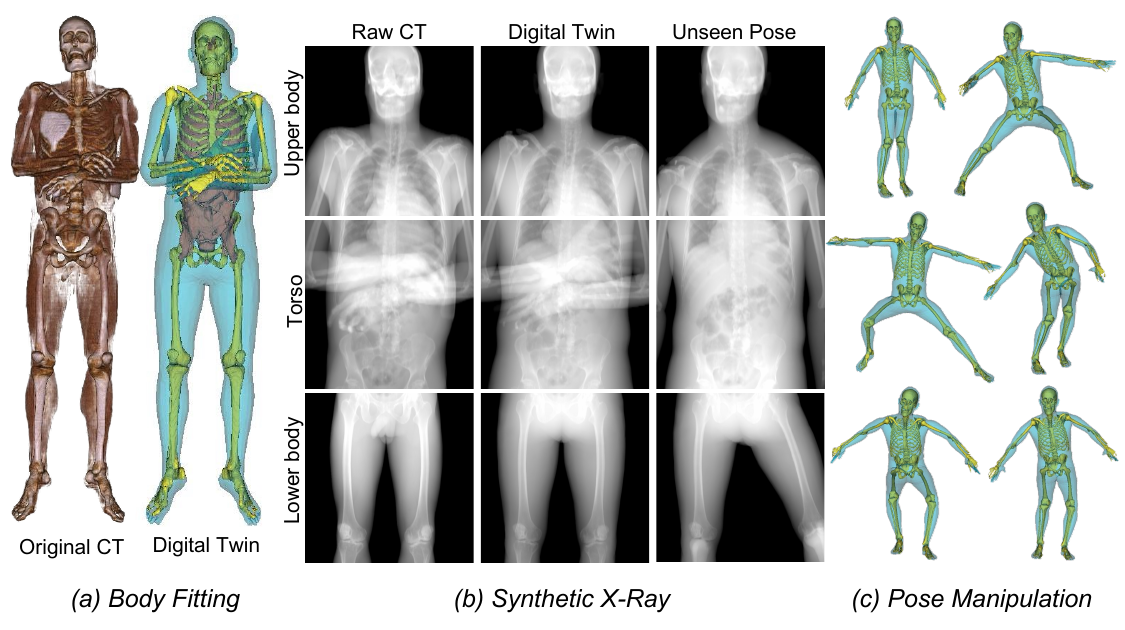}
    \caption{\textbf{(a) Body fitting:} the fitted body envelope encloses the CT-derived skeletal anatomy, shown alongside the original CT. \textbf{(b) Synthetic X-ray:} DRRs rendered from the original CT, the recomposed digital twin at the acquisition pose, and the articulated twin under an unseen target pose. \textbf{(c) Pose manipulation}: examples of the patient-specific twin articulated into diverse target poses.}     
    \label{fig:recon}
\end{figure}

\subsection{Body fitting and skeletal enclosure}
We first evaluated how well the fitted SMPL model represents the patient body in the CT acquisition pose. We measured geometric agreement between the fitted SMPL body surface and the CT-derived body surface using bidirectional chamfer distance. Since the CT acquisition pose differs from the canonical pose assumed by SMPL and the parametric body surface may not fully capture patient-specific shape, we also evaluated skeletal enclosure, defined as the fraction of CT-derived skeleton located inside the fitted SMPL. Both metrics were computed for each body region, partitioned along the cranio-caudal axis, and for the full body.

As shown in Table~\ref{tab_results}, the fitted body model achieved an overall chamfer distance of $15.8 \pm 4.0$ mm and enclosed $95.9\pm 1.8\%$ of the CT-derived skeleton. Fitting error was lowest for the upper body ($15.7 \pm 2.9$ mm) and highest for the lower body ($16.5 \pm 4.8$ mm), while skeletal enclosure remained high across regions, ranging from $94.1\%$ to $99.0\%$. 

The chamfer residual primarily reflects body-envelope agreement rather than anatomical registration error. Two factors dominate it: the parametric SMPL surface cannot represent patient-specific soft-tissue distribution, and the supine acquisition introduces deformation of the external surface that lies outside the SMPL shape space. Since the envelope acts only as a kinematic carrier and is not used to define bone geometry, enclosure is the operative criterion for subsequent motion retargeting; the high enclosure indicates that the fitted model provides a patient-aligned scaffold containing the underlying skeletal anatomy. Both metrics are reported per cranio-caudal region, and enclosure is defined on skeletal geometry only; organ placement is not evaluated in this work.

\begin{table}
\centering
\caption{Body-fitting geometry (Chamfer distance, enclosure) and recomposition fidelity (SSIM, PSNR).}
\label{tab_results}
\begin{tabular}{lcccc}
\toprule
& \multicolumn{2}{c}{Body fitting} & \multicolumn{2}{c}{Recomposition fidelity} \\
\cmidrule(lr){2-3}\cmidrule(lr){4-5}
Region & Chamfer (mm) $\downarrow$ & Enclosure (\%) $\uparrow$  & SSIM $\uparrow$ & PSNR (dB) $\uparrow$\\
\midrule
Upper body  & 15.7 $\pm$ 2.9 & 95.0 $\pm$ 2.4 & 0.890 $\pm$ 0.007 & 19.6 $\pm$ 1.3\\
Torso & 16.2 $\pm$ 3.5 & 94.1 $\pm$ 2.4 & 0.872 $\pm$ 0.026 & 17.5 $\pm$ 1.0\\
Lower body & 16.5 $\pm$ 4.8 & 99.0 $\pm$ 0.4 & 0.855 $\pm$ 0.035 & 18.4 $\pm$ 2.6\\
\midrule
Overall    & 15.8 $\pm$ 4.0 & 95.9 $\pm$ 1.8 & 0.872 $\pm$ 0.016 & 18.5 $\pm$ 1.4\\
\bottomrule
\end{tabular}
\end{table}

\subsection{Recomposition fidelity}
We next evaluated whether the recomposed patient twin preserves radiographic structure at the original CT acquisition pose, isolating the effect of kinematic binding and recomposition from that of pose change. We evaluated three anatomical regions: the upper body, torso, and lower body. For each region, eight projection views were rendered at 45-degree increments from both the original CT and the recomposed patient twin under identical imaging geometries, resulting in 72 paired DRR comparisons across the three subjects. The resulting DRRs were compared using SSIM and PSNR.

As summarized in Table~\ref{tab_results}, the recomposed twin achieved an overall SSIM of 0.872 $\pm$ 0.016 and PSNR of 18.5 $\pm$ 1.4 dB across all paired DRR comparisons. SSIM was highest for the upper body (0.890 $\pm$ 0.007) and lowest for the lower body (0.855 $\pm$ 0.035), while PSNR was highest for the upper body (19.6 $\pm$ 1.3 dB) and lowest for the torso (17.5 $\pm$ 1.0 dB). The residual error is mainly attributable to the constant-HU soft-tissue fill, which does not reproduce the heterogeneous attenuation present in the original CT. This intensity-level discrepancy depresses PSNR, which is sensitive to absolute per-pixel differences, more than SSIM, which emphasizes structural agreement, indicating that the major skeletal structures are well preserved despite the modest PSNR. Representative examples are shown in Fig.~\ref{fig:recon}(b). Visually, the recomposed DRRs match reference DRRs rendered from the original CT in bony anatomy and overall body contour, with residual differences mainly around distal extremities. 

\subsection{Unseen-pose articulation and structural consistency}
Finally, we evaluated structural consistency under unseen target poses. We sampled 28 representative AMASS motion sequences~\cite{mahmood_amass_2019}, spanning a range of articulations such as raised arms, flexed limbs, and trunk bending. Target poses were specified as SMPL pose parameters and retargeted to the CT-derived anatomy. Because ground-truth patient anatomy is not available for these reposed configurations, we do not directly evaluate biomechanical plausibility. Instead, we assess structural self-consistency by measuring enclosure between the reposed skeletal anatomy $\mathcal{M}^B_{\mathrm{bone}}$ and the reposed body envelope $\mathcal{M}^B_{\mathrm{SMPL}}$, and qualitatively demonstrate that the resulting geometry can be rendered as pose-dependent DRRs. Across the sampled poses, enclosure remained at 94.4 $\pm$ 0.4\%, comparable to the 95.9\% measured at the acquisition pose, indicating that articulation does not induce catastrophic skeleton–envelope penetration or rig breakdown. However, enclosure is a necessary but not sufficient criterion: a bone can remain inside the envelope while being incorrectly rotated or displaced relative to its neighbors. 

Fig.~\ref{fig:recon}(c) qualitatively shows the twin articulated into target poses with no obvious global misalignment, with the patient-specific skeletal geometry retained across configurations. Moreover, the synthetic radiographs exhibit pose-dependent changes in projection appearance (Fig.~\ref{fig:recon}(b)). Since no ground-truth radiographs exist for these target poses, we present these renderings as a qualitative feasibility demonstration of pose-diverse DRR generation from a single static CT.

\section{Discussion and Conclusion}
We presented a proof-of-concept for constructing a patient-specific articulated digital twin from a single full-body CT scan. The key idea is to use SMPL as a patient-aligned kinematic prior that enables CT-derived skeletal and organ geometry to be reposed into new body configurations. This bridges a practical gap in CT-based simulation: conventional DRR rendering can vary the imaging viewpoint, but the patient anatomy remains fixed in the acquisition pose. In contrast, our approach preserves CT-derived skeletal geometry while enabling controlled body-pose changes. In our evaluation, the fitted body envelope enclosed most of the skeletal anatomy and supported articulation into unseen poses, while recomposed DRRs preserved the major skeletal structures and body outline at the acquisition pose. Rendering the reposed anatomy as DRRs further illustrates the feasibility of pose-dependent radiographic simulation from a single static CT.

The current approach has several limitations. First, the evidence is limited in scale: all results are obtained from three subjects, and articulated configurations are assessed using enclosure and qualitative inspection rather than validated against ground-truth anatomy or radiographs. Second, the method primarily models skeletal articulation. Organs and soft-tissue structures are transformed using coarse, region-level rigid motions, so organ-specific non-rigid deformation, sliding interfaces, and gravity-dependent displacement are not modeled, and the residual soft tissue is filled at a constant HU that removes heterogeneous attenuation. Third, although SMPL provides a useful kinematic prior, it remains a parametric surface model: joint limits are inherited from the body model rather than from the patient's own articular surfaces, patient-specific body shape is only partially captured, and local mismatch or self-intersection may therefore arise under extreme poses.

We therefore position the resulting twins as a proof-of-concept platform for exploring pose-diverse radiographic simulation in algorithm development and positioning studies. Future work will evaluate the approach in larger and more diverse cohorts, validate articulated configurations using ground truth or physical phantoms, and assess performance in downstream image-guided intervention tasks.

\subsubsection*{Disclosure of Interests} The authors have no competing interests to declare that are relevant to the content of this article.

\begin{credits}
\subsubsection{\ackname} We acknowledge support from the National Science Foundation under Award No. 2239077, and Johns Hopkins University Internal Funds.
\end{credits}
\bibliographystyle{splncs04}
\bibliography{sn-bibliography}
\end{document}